\documentclass[letterpaper, 10 pt, conference]{ieeeconf}  

\IEEEoverridecommandlockouts                              

\usepackage{graphicx}
\usepackage{amsmath,amssymb,amsfonts}
\usepackage{algorithmic}
\usepackage{algorithm}
\usepackage{xcolor}
\usepackage{animate}
\usepackage{tabularx,booktabs}
\newcolumntype{Y}{>{\centering\arraybackslash}X}
\usepackage{subcaption}
\usepackage{multirow}

\title{\LARGE \bf
MINT-RVAE: Multi-Cues Intention Prediction of Human-Robot Interaction using Human Pose and Emotion Information from RGB-only Camera Data
}

\author{Farida Mohsen$^{1}$, Ali Safa$^{1}$
\thanks{$^{1}$ College of Science and Engineering, Hamad Bin Khalifa University, Doha, Qatar.}%
\thanks{F. Mohsen carried the technical developments, the data collection and the design of the experiments. A. Safa supervised the project as Principal Investigator and contributed to the technical and experimental planning. All authors contributed to the writing of the manuscript.
        {\tt\small fmohsen@hbku.edu.qa, asafa@hbku.edu.qa}}%
}

\begin{document}

\maketitle
\thispagestyle{empty}
\pagestyle{empty}


\begin{abstract}  
Efficiently detecting human intent to interact with ubiquitous robots is crucial for effective human-robot interaction (HRI) and collaboration. Over the past decade, deep learning has gained traction in this field, with most existing approaches relying on multimodal inputs, such as RGB combined with depth (RGB-D), to classify time-sequence windows of sensory data as interactive or non-interactive. In contrast, we propose a novel RGB-only pipeline for predicting human interaction intent with frame-level precision, enabling faster robot responses and improved service quality. A key challenge in intent prediction is the class imbalance inherent in real-world HRI datasets, which can hinder the model's training and generalization. To address this, we introduce MINT-RVAE, a synthetic sequence generation method, along with new loss functions and training strategies that enhance generalization on out-of-sample data. Our approach achieves state-of-the-art performance (AUROC: 0.95) outperforming prior works (AUROC: 0.90–0.91.2), while requiring only RGB input and supporting precise frame onset prediction. Finally, to support future research, we openly release our new dataset with frame-level labeling of human interaction intent.

\end{abstract}

\section*{Supplementary material}
The dataset used in this work, as well as the companion Appendix files, are openly available at: 
\texttt{https://tinyurl.com/2ymx6ahr}

\section{Introduction}
Service robots operating in public spaces should be able to recognize when humans intend to interact with them to respond in a timely and socially appropriate manner~\cite{lee2010receptionist}. In domains such as hotels, shopping centers, and healthcare facilities, this capability is essential for delivering seamless user experiences; for example, a robot receptionist must recognize that a person approaches with the intent to engage before explicit verbal or gestural signals occur~\cite{lidard2024risk}. Accurate intent detection improves fluency, safety, and user trust by minimizing delays and avoiding inappropriate or missed responses~\cite{yang2022model, valls2024robot}.  

\begin{figure}[t]
    \centering
    \includegraphics[width=0.49\textwidth]{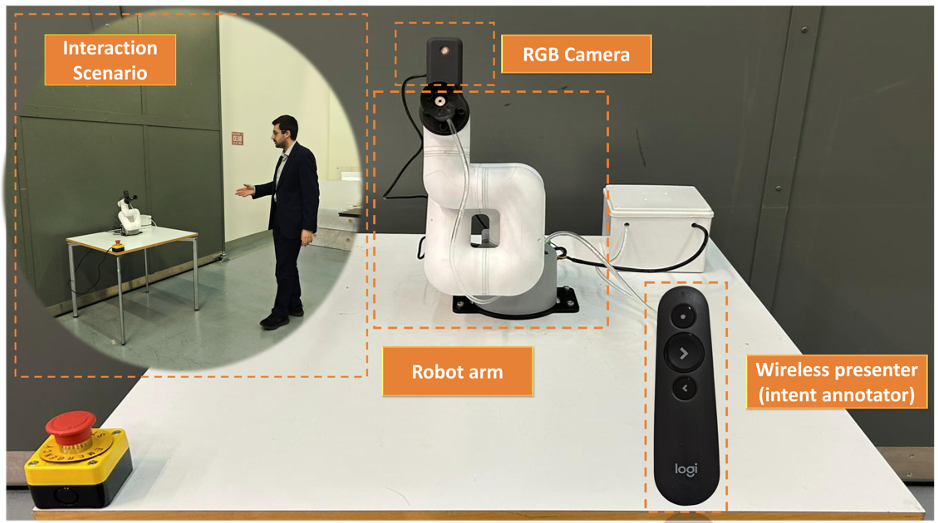}
  \caption{\textit{\textbf{Detecting the intention of humans to interact with a robot arm using RGB data only.} A monocular RGB camera mounted on an Elephant Robotics MyCobot 320 arm is used to detect human intent to interact with the robot. During data collection, subjects press a wireless presenter button at the moment they intend to interact, providing frame-accurate annotations synchronized with the video. A GPU-equipped laptop runs real-time YOLOv8-based human detection and pose extraction \cite{ultralytics_yolov8}, along with the DeepFace emotion detection model~\cite{serengil2020deepface}, yielding multimodal pose and emotion data for intent prediction.
  }}
  \vspace{-1em}
   \label{fig:graphabstr}
\end{figure}

\begin{figure*}[htbp]
    \centering
    \begin{subfigure}[t]{0.48\linewidth}
        \centering
        \includegraphics[width=0.7\linewidth]{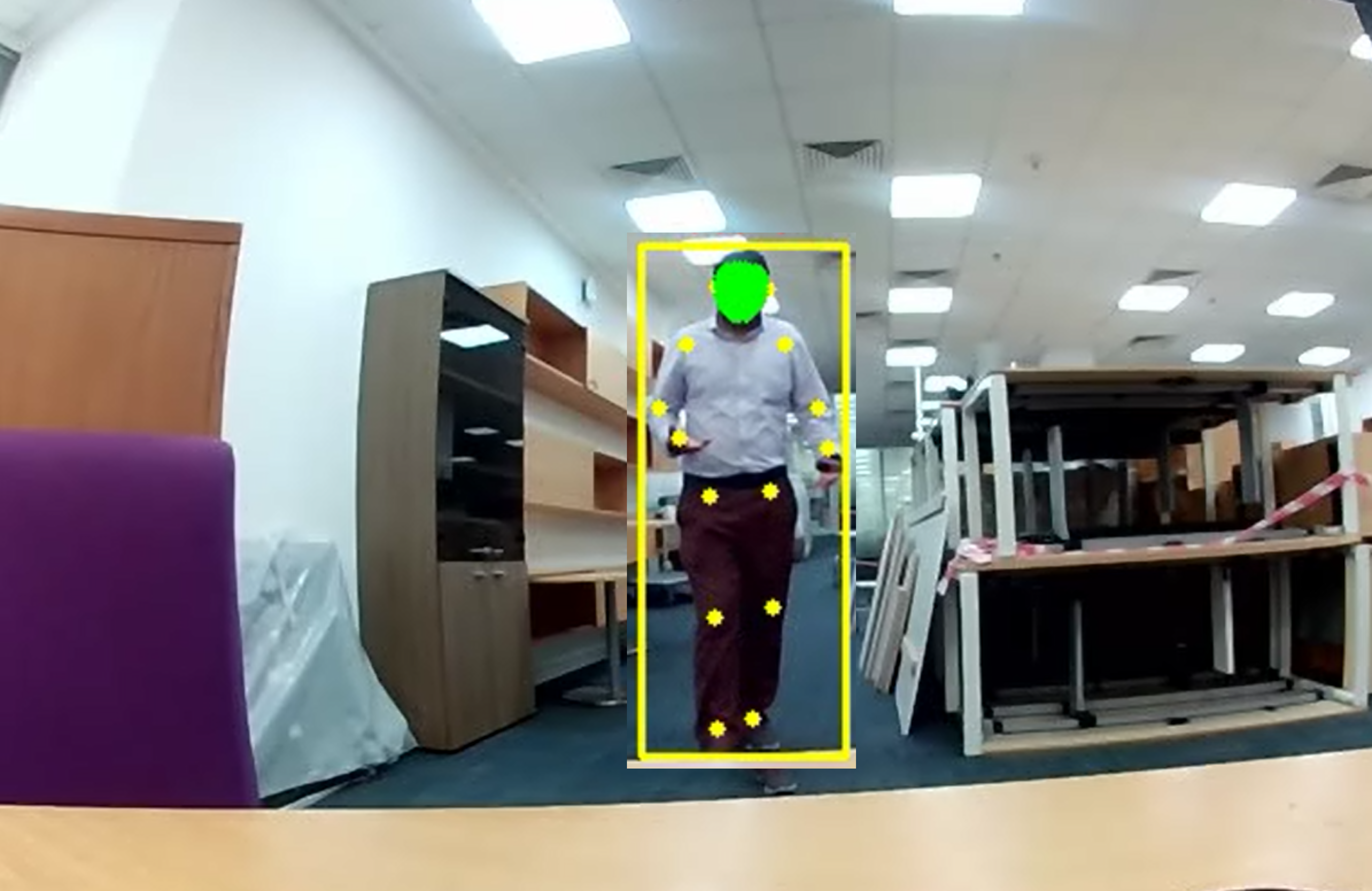}
        \label{fig:one_person}
    \end{subfigure}
    \hfill
    \begin{subfigure}[t]{0.48\linewidth}
        \centering
        \includegraphics[width=0.7\linewidth]{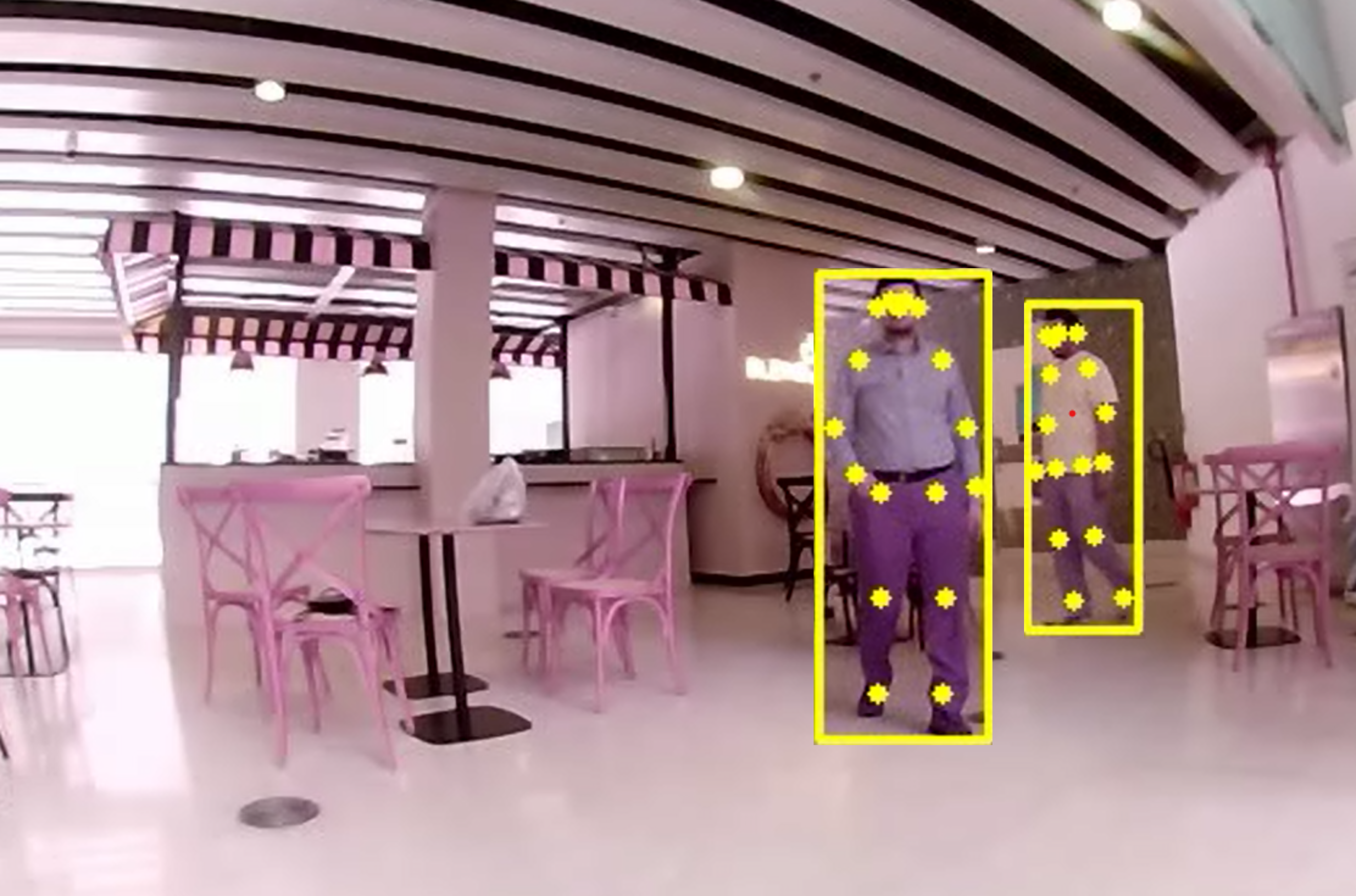}
        \label{fig:2_person}
    \end{subfigure}
\caption{\textit{\textbf{Views from our data collection and processing procedure in both the single-person and multi-person settings.} A YOLOv8-pose model \cite{ultralytics_yolov8} is used for both detecting the human subjects and for retrieving the 2D human pose coordinates together with their confidence indicators. Furthermore, a DeepFace model \cite{serengil2020deepface} is used to retrieve a categorical emotion distribution vector from the human facial analysis.}}
    \label{fig:snapshoots}
    \vspace{-1em}
\end{figure*}

Recent research on human–robot interaction (HRI) has focused on human intent prediction for collaborative tasks between humans and robots \cite{kedia2024interact}. These models, typically trained on relations between human motion trajectories and inferred intentions, enable robots to adapt or plan their actions proactively. However, many existing methods rely on more specialized hardware, such as \textit{depth cameras} or \textit{motion capture systems} (MOCAP), to capture 3D poses~\cite{kedia2024interact, arreghini2024predicting}, limiting system scalability and leading to more significant equipment costs that are less suited for real-world robot system deployments~\cite{gaschler2012social, belardinelli2022intention}. Moreover, the high \textit{variability} of human nonverbal cues constitutes one of the main challenges for the robust detection of human interaction intent, where trained models must \textit{generalize well} to out-of-distribution test data unseen during training. Humans rely on posture, facial expressions, and proxemics to signal social intent~\cite{saunderson2019robots}, but endowing robots with the ability to interpret such signals requires multimodal perception and temporal reasoning~\cite{gasteiger2021factors}. 
 
Another limitation is the frame labeling resolution. Three common practices are observed: \textit{i)} Sequence-level labels: most studies provide intent labels for entire data segments, such that all frames within a temporal window are assigned to the same intent label \cite{arreghini2024predicting,abbate2024self}; \textit{ii)} Action segmentation post-processing: some approaches assign intent labels after segmenting actions or interactions in post-processing, with intent annotation occurring retrospectively based on observed events \cite{trick2023can,hernandez2024bayesian}; \textit{iii)} Frame-wise outputs with sequence labels (coarse evaluation): other studies produce a probability per frame but train and evaluate their model using the \textit{same} label replicated across frames, that is, without frame-accurate \textit{onset} annotation  \cite{kedia2024interact,arreghini2024predicting,abbate2024self}. While suitable for sequence classification, such methods omit the critical moment when intention first emerges (i.e., the \textit{onset}) and hence, jeopardizes the system's response time. More broadly, surveys on human motion and intent prediction stress that temporal fidelity is critical for responsive behavior: earlier, more precise predictions enable earlier actions \cite{rudenko2019human}. Recognizing this onset at the frame level (vs. sequence level) is crucial for practical service robots, as it determines the earliest point at which a robot can and should act, thereby leading to an elevated quality of service and robot responsiveness.

\begin{figure*}[t]
    \centering
    \includegraphics[width=0.9\textwidth]{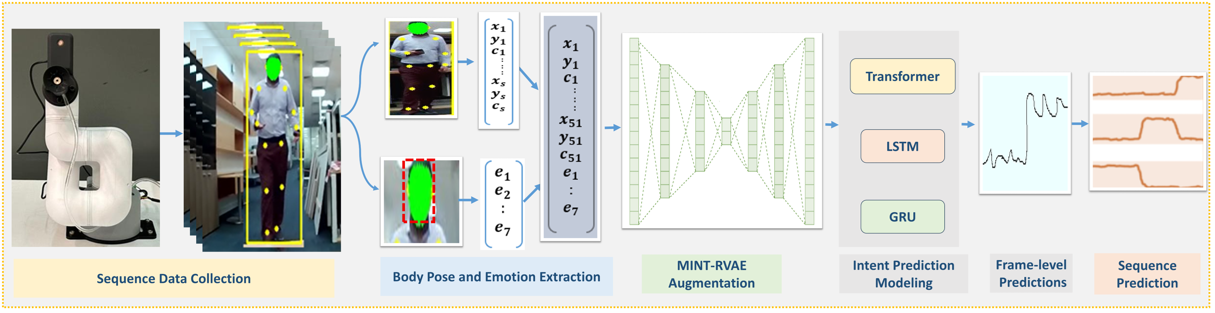}
  \caption{\textit{\textbf{Proposed data processing pipeline.} The RGB camera feed from the robot arm is processed using a \texttt{YOLOv8-pose} model for pose coordinates and \texttt{DeepFace} for emotion vectors. These outputs are concatenated into multimodal feature vectors, which serve as input to various intent detection backbones studied in this work (\texttt{GRU}, \texttt{LSTM}, \texttt{Transformer}). To address the class imbalance in HRI data, our proposed \textit{MINT-RVAE} generates synthetic sequences during training. The pipeline supports both frame-level and sequence-level intent prediction.
  } }
   \label{fig:study_diagram}
   \vspace{-1em}
\end{figure*}
Another central challenge in HRI intent prediction is \textit{class imbalance}. In public deployments, genuine interaction events are rare relative to prolonged non-interaction; for example, the PAR-D dataset in \cite{thompson2024pard} reports 112 / 4245 interacted vs.\ non-interacted trajectories. Furthermore, field studies in shopping malls observe engagement rates of \(\sim 3.6\%\) \cite{natori2025mall}. Imbalance manifests at both the \textit{sequence} level (few sequences contain interactions) and the \textit{frame} level (positive labels are vastly outnumbered), motivating \textit{imbalance-aware training} and imbalance-robust metrics (AUROC, macro-\(F_{1}\), and balanced accuracy). Common remedies such as \textit{undersampling} \cite{del2020still} or synthetic oversampling with SMOTE-like methods \cite{chawla2002smote} have limitations in sequential, multimodal settings: undersampling reduces effective data and discards informative negatives, while SMOTE-like oversampling disrupts temporal coherence and multimodal consistency due to their \textit{non-learning-based} approach (vs. generative models such as variational autoencoders \cite{kingma2014autoencoding} and generative adversarial networks \cite{goodfellow2020generative}), ultimately degrading the performance of sequential models. 

To address the above limitations and challenges in HRI intent detection contexts, this paper proposes a novel \textit{RGB-only} pipeline for accurate \textit{frame-level} detection of human intentions to interact with a robot arm (see Fig. \ref{fig:graphabstr}). Unlike prior works that mostly focus on \textit{sequence-level} detection and rely on more expensive sensor suites  (e.g., RGB--D) \cite{arreghini2024predicting,abbate2024self},  our approach reduces hardware cost (approximately \(\$350\) for an Intel RealSense RGB--D camera vs.\ \(\$10\) for a commodity USB webcam), while enabling onset-accurate decisions. To tackle class imbalance, we introduce \emph{MINT--RVAE}, a multimodal recurrent variational autoencoder (VAE) \cite{kingma2014autoencoding} for imbalance-aware sequence augmentation, which significantly improves the generalization of intent detector models. The contributions of this paper are as follows:
\begin{enumerate}
    \item \emph{Dataset:} We release a novel dataset (anonymized) with \emph{frame-level onset annotations} of human intention to interact with a robot arm across diverse indoor public-space scenarios, collected with informed consent and privacy safeguards. To our knowledge, this is the first publicly available dataset that enables the study of frame-level intent prediction, in contrast to previous datasets that provide only sequence-level labels.       
    
    \item \emph{MINT-VAE for HRI data re-balancing:} We introduce a multimodal recurrent VAE together with custom loss function combinations and training strategies for learning a joint sequential latent representation over pose, emotions, and intent labels, with the goal of generating \emph{temporally coherent, cross-modal} synthetic sequences for alleviating training data imbalance

    \item \emph{RGB-only intent detection:} We develop a novel modular pipeline for the detection of human interaction intent (i.e., performing detection before any explicit interaction takes place) using only a single RGB camera, substantially simplifying hardware and cost overheads.%

    \item \emph{Experiments:} We train three different backbone networks (\texttt{GRU}, \texttt{LSTM}, \texttt{Transformer}), while using our proposed MINT-RVAE for data augmentation, and show that the resulting pipeline achieves strong detection performance in both frame-level and sequence-level cases (AUROC: $0.95$), outperforming previously introduced methods (with typical AUROC: $0.90$-$0.92$). 
\end{enumerate}
The remainder of this paper is organized as follows. First, our dataset collection and processing pipeline are described in Section \ref{dataarch}. The proposed \textit{MINT-RVAE} method for HRI data rebalancing is detailed in Section \ref{mintrvae}. The experimental results are discussed in Section \ref{resultssection}. Finally, conclusions are provided in Section \ref{concsection}.

\section{Data processing architecture}
\label{dataarch}


This section first introduces the HRI dataset we collected for frame-level intent prediction and then outlines our modular data-processing architecture composed of several deep learning modules (Fig.~\ref{fig:study_diagram}).

\subsection{Dataset collection}
\label{datacolllpre}


Using the setup in Fig.~\ref{fig:graphabstr}, we collected a diverse RGB video dataset for human–robot interaction intent detection across three different indoor environments with $10$ different participants. The dataset provides \emph{frame-level} intent labels together with human pose coordinates and facial emotion vectors. The recordings involved participants performing a diverse set of approach behaviors while labeling their interaction intent using the wireless presenter (see Fig.~\ref{fig:graphabstr}). All participants agreed to participate and to the use of the data. To preserve anonymity, we did not collect personally identifying information or demographics. The released dataset contains only de-identified extracted features (e.g., 2D pose key points). Table~\ref{tab:dataset_summary} summarizes the dataset. In addition, Fig. ~\ref {fig:snapshoots} shows example views from the collection and processing pipeline.

\begin{table}[t]
\centering
\caption{\textit{\textbf{Overview of our collected dataset.} Entries show the number of videos and total frames collected at each environment; the percentage in parentheses is the number of frames labeled as interaction intent.}}
\setlength{\tabcolsep}{8pt}
\begin{tabular}{lcc}
\toprule
\textbf{Environment} & \textbf{Sequences} & \textbf{Frames (\,\% intent\,)} \\
\midrule
Library (Env.~1)   & 54 & 7{,}620 (30.2\%) \\
Corridor (Env.~2)  & 23 & 3{,}900 (32.8\%) \\
Two-person (Env.~3) & 11 & 1{,}095 (37.5\%) \\
\bottomrule
\vspace{-2em}
\end{tabular}
\label{tab:dataset_summary}
\end{table}

The dataset is organized into a collection of sequences. 
Let $S$ be the number of sequences, sequence $j$ contains $k=1,...,N_j$ frames with associated feature vectors $f_{k,j}$ and binary frame-level labels $y_{k,j}$: $\{(f_{k,j},y_{k,j})\}_{k=1}^{N_j}$, where $y_{k,j}\in\{0,1\}$ indicates \emph{intent} (1) vs.\ \emph{no intent} (0).
\noindent \begin{equation}
D=\{(f_{k,j},y_{k,j})\mid j=1,\dots,S;\; k=1,\dots,N_j\}
    \label{datasetdescr}
\end{equation}

Each feature vector $f_{k,j}$ concatenates the human pose and emotion modalities:
\[
f_{k,j}=\big[f^{\text{pose}}_{k,j};\, f^{\text{emo}}_{k,j}\big].
\]

\subsubsection{Pose feature vector and pre-processing} Let $K$ denote the number of body keypoints (joints) that consist of 2D pixel coordinates detected on the camera image plane. We use a \textit{YOLOv8-pose} model~\cite{ultralytics_yolov8} to extract 2D pixel coordinates $(x_m,y_m)$ and detector confidences $c_m$ for each pose keypoint $m=1,...,M$. 
Extracting pose information as input data to our intent detection models is important since it reduces the raw camera data into pose sequences that provide a direct indication of the user’s body language. In addition to the pose coordinates, the bounding box coordinates $b_{k,j}=(x_{\min},y_{\min},w,h)$ indicating the detected person is also returned as well, where $x_{\min},y_{\min}$ denote the location of the top-left bounding box corner and $w,h$ denote the bounding box width and height.

To provide invariance to scale and translation across scenes, we normalize all pose coordinates with regard to the bounding box coordinates as follows:
\[
({x}_m,{y}_m)=\Big(\tfrac{x_m-x_{\min}}{w},\,\tfrac{y_m-y_{\min}}{h}\Big),
\]
Doing so, we obtain normalized pose feature vectors:
\[
f^{\text{pose}}_{k,j}\in\mathbb{R}^{3M},\qquad
f^{\text{pose}}_{k,j}=[\,{x}_1,{y}_1,c_1,\,\dots,\,{x}_M,{y}_M,c_M\,].
\]

Finally, we standardize the $2M$ box–normalized coordinates using standard $z$–normalization \cite{9710829} using \emph{training-set} statistics (mean and standard deviation), leaving all confidences $c_m$ unchanged (not $z$-normalized).

\subsubsection{Face detection and emotion probabilities} we again make use of a YOLOv8 model for detecting human faces ~\cite{ultralytics_yolov8}. After face detection and face region cropping, we use a \textit{DeepFace} model~\cite{serengil2020deepface} for obtaining an emotion probability vector denoting the probability of seven different types of standard human emotions (happy, sad, angry, etc.). Doing so, we obtain an emotion feature vector noted as:
\[
f^{\text{emo}}_{k,j}\in\mathbb{R}^{7},\hspace{3pt}
f^{\text{emo}}_{k,j}=[p_1,\dots,p_7],\hspace{3pt}
p_m\in[0,1],\ \sum_{m=1}^{7} p_m=1.
\]

\subsection{Alleviating the naturally-present data imbalances}


As summarized in Table~\ref{tab:dataset_summary}, the collected dataset is imbalanced between \emph{intent} and \emph{no-intent} (only \(\sim\!30\%\) of frames are labeled as \emph{intent}), a pattern typical of public deployments (e.g., in retail). Such imbalance can degrade training and hinder the \textit{generalization} of the trained models on out-of-distribution data encountered in practice  \cite{chawla2002smote}. To mitigate this, we propose a generative approach that produces class-balanced, temporally coherent multimodal sequences closely matching the real distribution. Our method, termed \textit{MINT-RVAE}, is presented in the next Section and used to augment the minority class during training.

\section{MINT-RVAE: recurrent variational auto-encoding for data sequence re-balancing}
\label{mintrvae}

\begin{figure*}[t]
    \setlength{\abovecaptionskip}{0pt}   
    \setlength{\belowcaptionskip}{0pt}   
    \centering
    \includegraphics[width=0.99\linewidth]{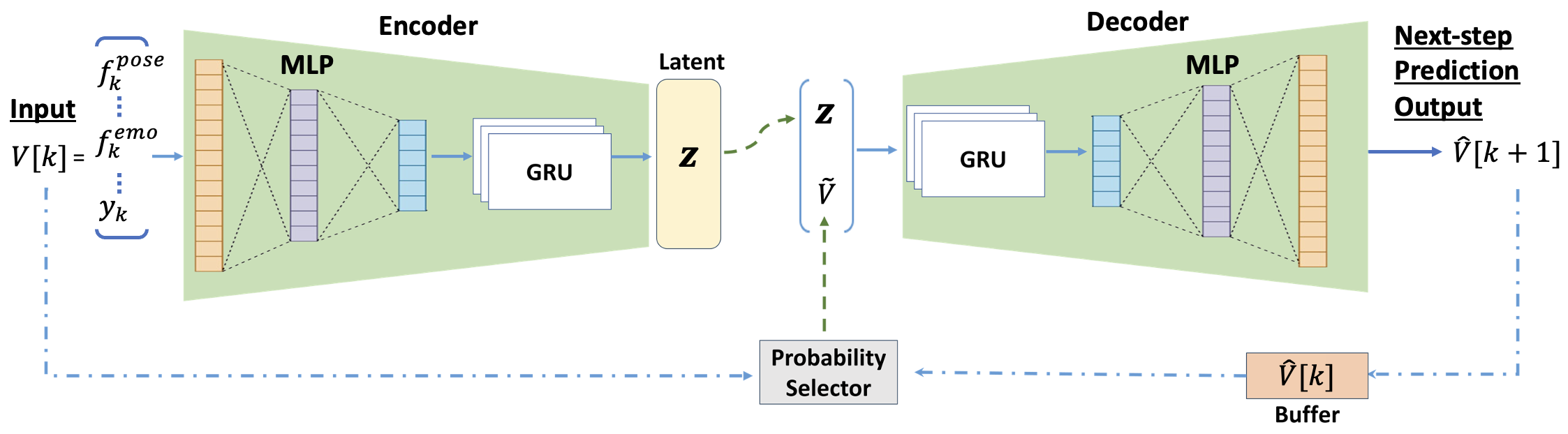}
    \caption{\textit{\textbf{Schematic overview of the proposed MINT-RVAE}. An input sequence \(x_{k=1:T}\) (each frame concatenating $f^{pose}_{k,j}$, $f^{emo}_{k,j}$ and $y_{k,j}$) is first processed by an MLP before being fed to an encoder GRU. The final encoder state parametrizes a latent \(z \in \mathbb{R}^{L}\) (block \(L\)). 
    At decoding time, \(z\) initializes the decoder GRU hidden state and is concatenated with the projected input at every step. The output head estimates the next frame (pose, emotion and label). The blue dashed path is a teacher forcing or autoregressive feedback that supplies the next decoder input. The Probability Selector block (used during training only) sets the teacher forcing probability. 
    At inference, \(z\) is sampled from  \(\mathcal{N}(0,I)\) and the selector always feeds back \(\hat{x}_k\).}}
    \label{fig:rvae_schematic}
    \vspace{-1em}
\end{figure*}

This Section describes \emph{MINT--RVAE}, a multimodal recurrent variational autoencoder that synthesizes realistic pose–emotion–label sequences to mitigate class imbalance in our training data (Table~\ref{tab:dataset_summary}). As illustrated in Fig.~\ref{fig:rvae_schematic}, the model jointly learns to model the temporal dynamics of body pose, facial affect, and interaction labels. Efficiently training MINT-VAE can be highly challenging due to the multimodal nature of its input data, with various data sources conveying various levels of information. If no care is taken, the MINT-VAE model can overfit on a single data source only, jeopardizing its \textit{multimodal} generative nature. In addition to this, we observed during our developments that the standard loss functions typically used during VAE training (e.g., the mean-square error) were \textit{not} leading to accurate input data reconstruction at the decoder's output, surely due to the specific nature of the pose and emotion vectors considered in this work (vs. typical image data, for example). Accordingly, we design custom loss function combinations and optimization schedules for yielding high quality synthetic data generation.


\subsection{Model architecture}
\subsubsection{Input sequences}
We design MINT-RVAE (see Fig. \ref{fig:rvae_schematic}) to jointly reconstruct pose dynamics, emotion probability, and intent labels from chunks of data sequence (with a window size of $T = 15$ frames).  
We note each sequence $j$ as:
\begin{equation}
  V_j = \{x_{1,j}, x_{2,j}, \dots, x_{T,j}\}, \hspace{3pt} V_j[k] = x_{k,j} \in \mathbb{R}^D,  
  \label{inputseq}
\end{equation}
where each multimodal data vector $x_{k,j}=\{f^{pose}_{k,j}, f^{emo}_{k,j}, y_{k,j} \}$ is the concatenation of \textit{1)} a 51-dimensional 2D body pose coordinates with an additional dimension denoting the confidence values on every pose coordinate $f^{pose}_{k,j}$, \textit{2)} a 7-dimensional emotion probabilities vector $f^{emo}_{k,j}$, and \textit{3)} a binary intent label $y_{k,j}$ (indicating the intention of \textit{"interaction"} or \textit{"no interaction"}), ensuring that the model learns to reproduce both behavioral signals and the associated intent annotations. This leads to concatenated data vectors $x_{k,j}$ of dimension $D = 59$.

\subsubsection{Encoder network}
The encoder $q_\phi(z|V)$ maps a multimodal input sequence (\ref{inputseq}) into an associated sequence of latent vectors $z$ distributed following a standard Gaussian distribution.
First, each multimodal vector $x_{k,j}$ within the sequence $V_j$ is processed through a feedforward \textit{Multilayer Perceptron} (MLP) comprising three layers of dimensions $256 \!\to\! 128\!\to\! 64$.  Each layer consists of a linear transformation followed by batch normalization, a ReLU nonlinearity, and dropout, which enable \textit{modality fusion} and the reduction of data dimensionality.  
The MLP output sequence is then processed by a \textit{Gated Recurrent Unit} (GRU) \cite{Chung2014GRUEval} to capture the temporal dependencies between the successive multimodal input vectors $x_{k,j}$.  
The recurrent hidden state of the GRU is then passed through a pair of linear layers $\mu_\gamma, \sigma_\rho$ in order to yield the parameters (mean and standard deviation) of the variational posterior:
\[
q_\phi(z_k|x_{k,j}) = \mathcal{N}\!\left(z_k;\; \mu_\gamma(x_{k,j}),\, \mathrm{diag}(\sigma^2_\rho(x_{k,j}))\right),
\]
with mean $\mu_\gamma(x_{k,j})$ and log-variance $\log\sigma^2_\rho(x_{k,j})$.  
A latent sample $z \in \mathbb{R}^k$ with size 32 is obtained using the \textit{reparameterization trick} \cite{kingma2014autoencoding}:
\[
z_k = \mu_\gamma(x_{k,j}) + \sigma_\rho(x_{k,j}) \odot \epsilon, 
\quad \epsilon \sim \mathcal{N}(0, I).
\]

\subsubsection{Decoder network} 
The decoder $p_\theta(V|z)$ consists of an \textit{autoregressive} GRU layer followed by an MLP, which seeks to predict the \textit{next multimodal vector} $\hat{x}_{k+1,j}$ from the previous GRU hidden state $h_{k}$ and the latent state $z_k$ that it received as input from the encoder (i.e., one-step-ahead prediction). 

The resulting GRU hidden state at time step $k+1$, $h_{k+1}$, is then processed through an output MLP back into the multimodal vector space of the input data, yielding an estimate of the next multimodal vector $\hat{x}_{k+1,j}$ within the sequence:
\begin{equation}
 \hat{x}_{k+1,j} = \mathrm{concat}(\hat{f}^{pose}_{k+1,j}, \hat{f}^{emo}_{k+1,j}, \hat{y}_{k+1}).   
\end{equation}
where $\hat{f}^{pose}_{k+1,j}$ are the reconstructed pose keypoints (linear output), $\hat{f}^{emo}_{k+1,j}$ is the softmax-processed emotion distribution vector, and $\hat{y}_{k+1} \in [0,1]$ is the sigmoid-processed interaction label.

\vspace{-0.2em}
\subsection{MINT-RVAE training approach}
\label{sec:training_objective}
For each data sequence window, we form an input sequence $V[k], k=1:T-1$ and a \emph{next-step} target sequence $V[k], k=2:T$ that the MINT-RVAE model needs to learn to predict. The MINT-RVAE recurrently encodes $V[k], k=1:T-1$ into a latent sequence $z_k$ and decodes this sequence back in an autoregressive manner in order to estimate the \textit{one-step-ahead} prediction of the multimodal input vector sequence $\hat{V}[k], k=2:T$ (see Fig. \ref{fig:rvae_schematic}).

As total loss function $\mathcal{L}_{\text{total}}$, we optimize the following multivariate loss:
\begin{equation}
\mathcal{L}_{\text{total}}
= \lambda_p \,\mathcal{L}_{\text{pose}}
+ \lambda_e \,\mathcal{L}_{\text{emo}}
+ \lambda_y \,\mathcal{L}_{\text{label}}
+ \beta_e \,\mathcal{L}_{\text{KL}},
\label{lossall}
\end{equation}
composed of \textit{i)} a loss term for the pose vector estimation $\mathcal{L}_{\text{pose}}$; \textit{ii)} a loss term for the emotion vector estimation $\mathcal{L}_{\text{emo}}$; \textit{iii)} a loss term for the HRI intention label estimation $\mathcal{L}_{\text{label}}$; and \textit{iv)} a standard KL divergence term $\mathcal{L}_{\text{KL}}$ between the latent space of MINT-RVAE and a standard Gaussian distribution \cite{kingma2014autoencoding}. Furthermore, each loss term is balanced through its associated hyperparameters $\lambda_p, \lambda_e, \lambda_y, \beta_e$.

\subsubsection{Pose loss}

We build a novel custom pose loss by combining a novel \textit{confidence-weighted} Huber loss (smooth-$\ell_1$) for the pose \textit{coordinates} together with an auxiliary mean-square error (MSE) loss for the pose \textit{confidence} indicators:
\begin{multline}
\mathcal{L}_{\text{pose}}
= 0.8 \cdot \frac{1}{T-1}\sum_{k=1}^{T}\sum_{j=1}^{J} (c_{k+1,j}+\nu)\,
\rho_\delta\!\big(\hat{p}_{k+1,j}-p_{k+1,j}\big)
\;
\\
+\;
0.2 \cdot \mathrm{MSE}(\hat{\mathbf{c}}_{k=2:T},\mathbf{c}_{k=2:T}),
\label{lposecomp}
\end{multline}
where the confidence metric $c_m$ associated to each pose coordinate $m$ weights the contribution of the Huber loss $\rho_\delta\!\big(\hat{p}_{k+1,j}-p_{k+1,j}\big)$ associated with this coordinate. In addition, we set $\nu{=}0.1$ to provide a minimum confidence value in cases where $c_m=0$. Our newly-proposed \textit{confidence-weighted Huber loss} in (\ref{lposecomp}) is motivated as follows. Adopting a Huber loss \cite{huberloss} instead of a standard MSE improves the model robustness against outliers that can originate from e.g., occlusion or motion blur \cite{girshick2015fast}. In addition, our custom weighting of the Huber loss by the pose coordinate confidence values \(\big(c_{k,j}+\nu\big)\) emphasizes the learning of the reliable joints while down-weighting the occluded and noisy ones. This improves the robustness to occlusions and to the pose detection noise, enhancing the fidelity of the output pose sequences synthetically generated by the model.  


The Huber loss is defined as \cite{huberloss}: 
\begin{equation}
\rho_\delta(r)=
\begin{cases}
\frac{1}{2\delta}\|r\|_2^2, & \|r\|_2 \le \delta,\\
\|r\|_2 - \frac{\delta}{2}, & \text{otherwise},
\end{cases}\qquad
\end{equation}
where $\delta = 1$ for $z$-normalized data (as in our normalized pose vectors). Furthermore, the $0.8$-$0.2$ balance in (\ref{lposecomp}) between the pose coordinate loss and confidence loss was tuned empirically during our model evaluation procedure (see Section \ref{modelvalid}).

\subsubsection{Emotion loss}
As the emotion loss, we make use of a KL divergence between the target categorical emotion distribution vector $\mathbf{e}_{k+1}$ and the estimated emotion vector $\hat{\mathbf{e}}_{k+1}$, which boils down to: 
\begin{equation}
\mathcal{L}_{\text{emo}}
=\frac{1}{T}\sum_{k=1}^{T}\sum_{c=1}^{C} e_{c,k+1}\,\log\frac{e_{c,k+1}}{\hat{e}_{c,k+1}}
\end{equation}


\subsubsection{Label loss}
As the interaction label prediction loss, we make use of a standard binary cross-entropy loss function, since the interaction labels result from a sigmoid function at the output of the MINT-RVAE decoder: $\mathcal{L}_{\text{label}}=BCE(y_{k+1},\hat{y}_{k+1})$.

\subsubsection{KL loss with free-bits regularization}
The sequence posterior is $q_\phi(z|V_{1:T})=\mathcal{N}(\boldsymbol{\mu},\mathrm{diag}(\boldsymbol{\sigma}^2))$
with prior $p(z)=\mathcal{N}(0,I)$. The standard KL loss between the latent space of MINT-RVAE and a standard Gaussian distribution boils down to: 
\begin{equation}
    \mathrm{KL}_m=\tfrac{1}{2}(\mu_m^2+\sigma_m^2-\log\sigma_m^2-1)
\end{equation}
for each entry $m$ of the latent vector. To further mitigate \textit{posterior collapse} during training, we clamp the KL divergence values using the \textit{free-bits} floor method \cite{10.5555/3157382.3157627} in order to obtain our final KL loss $\mathcal{L}_{\text{KL}}$:
\begin{equation}
\begin{aligned}
\mathcal{L}_{\text{KL}} &= \sum_{q=1}^{K} \max\!\big(\mathrm{KL}_q,\ \delta_{\mathrm{FB}}\big), \hspace{5pt}
\delta_{\mathrm{FB}} = 0.1
\end{aligned}
\end{equation}

During training, we adopt a linear warm-up approach for the importance weight $\beta_e$ associated with $\mathcal{L}_{\text{KL}}$ in (\ref{lossall}) throughout the training epochs $e$:
\begin{equation}
\begin{aligned}
\beta_e &= \beta_{\max}\cdot \min\!\left(\frac{e}{E_{\mathrm{warm}}},\,1\right),
\end{aligned}
\end{equation}
with $\beta_{\max} = 0.8, E_{\mathrm{warm}}=5000$. We also use \emph{teacher forcing} and \emph{scheduled sampling} for one–step–ahead decoding~\cite{WilliamsZipser1989,Bengio2015ScheduledSampling}. At step $k$, the decoder input $\tilde{x}_k$ is either the ground–truth frame $x_{k}$ (teacher forcing) or the model’s previous prediction $\hat{x}_k$ (autoregressive):
\begin{equation}
\tilde{x}_k =
\begin{cases}
x_k, & \text{with prob.} \hspace{3pt} \tau \\
\hat{x}_k, & \text{with prob.} \hspace{3pt} 1-\tau \, ,
\end{cases}
\end{equation}
where the teacher–forcing probability $\tau\!\in\![0,1]$ is annealed \emph{linearly} from $1$ to $0$ during training. Hence, early epochs condition the learning on ground truth inputs for stability, while later epochs rely increasingly on model outputs. This yields a fully autoregressive decoder at inference time seeded by the initial frame $x_{0,j}$ to generate $\hat{x}_{1,j},\hat{x}_{2,j},\dots$.

We set \(\lambda_p=20\), \(\lambda_e=10\), and \(\lambda_y=1\) in (\ref{lossall}) based on empirical tuning to optimize the realism of generated sequences. We train with Adam \cite{kingma2017adammethodstochasticoptimization} (learning rate \(10^{-3}\), \(L_{2}\) weight decay \(10^{-5}\)), using a batch size of 64 for 700 epochs. Fig. A1 in the Appendix shows the loss evolution during training.
\vspace{-0.3em}
\subsection{MINT-RVAE model validation}
\label{modelvalid}

After training, we sample a latent \(\mathbf{z}\sim\mathcal{N}(\mathbf{0},\mathbf{I})\) and decode in an autoregressive manner to generate multimodal sequences (pose, emotion, and intent labels) used for downstream augmentation. To quantify the quality of the generated synthetic sequences, we train a lightweight RNN to classify \emph{real} vs.\ \emph{synthetic} sequences (using an 80\%-20\% train-test split), and we report the test accuracy \(\mathrm{acc}\) and the discriminative score \(D=\lvert 0.5-\mathrm{acc}\rvert\), following standard protocols for synthetic time-series evaluation \cite{yoon2019timegan,pei2021towards}. Lower \(D\) indicates greater sequence realism. Our model yields \(D=0.077\) with \(\mathrm{accuracy}=0.577\), indicating a high level of synthetic sequence realism. Qualitatively, embeddings of generated and real sequences substantially overlap when visualized using t-SNE in Fig. A2 in the Appendix. Accordingly, the next section employs \emph{MINT--RVAE} to rebalance the training data for our intent detectors. 

\section{Experimental Results}
\label{resultssection}
This Section first outline our evaluation setup for intent detection model assessment. Then, we report the experimental results, as well as the ablation studies on the use of our proposed multimodal pipeline and \textit{MINT-RVAE} approach.

\subsection{Setup}

\subsubsection{Interaction prediction models}

In this work, we assess the effect of our proposed MINT-RVAE synthetic data re-balancing approach using three temporal backbones: \texttt{GRU}, \texttt{LSTM}, and a lightweight \texttt{Transformer} for detecting the intention of human subjects for interacting with the robot arm. The Transformer is a compact encoder with a learnable positional embedding, one encoder block, four heads, and a LayerNorm + linear head. We also use single-layer \emph{LSTM} and \emph{GRU} models with a linear classification head. For all models, we tune the hidden size: \emph{pose-only:} \(H=256\), and \emph{emotion-only:} \(H=16\). For \emph{pose+emotion}, we evaluate \(H \in \{96,128,256\}\) and report the best setting, \(H=96\) for the GRU and LSTM, and \(H=256\) for the transformer.


\begin{table*}[t]
\centering
\caption{Five-fold cross-validation performance metrics for the detection of human intentions to interact with the robot (mean \(\pm\) s.d.). Ablations over input configuration and MINT--RVAE rebalancing; best values in \textbf{bold}.}

\label{tab:id}
\setlength{\tabcolsep}{3.5pt}
\begin{tabular}{l l c c c c c c}
\toprule
\textbf{Architecture} & \textbf{Variant} & \textbf{Frame F1 (Macro)} & \textbf{Frame Bal. Acc.} & \textbf{Frame AUROC} & \textbf{Seq F1 (Macro)} & \textbf{Seq Bal. Acc.} & \textbf{Seq AUROC} \\
\midrule
\multirow{3}{*}{GRU}
& Pose-only
& 0.814 $\pm$ 0.056 & 0.809 $\pm$ 0.051 & 0.899 $\pm$ 0.066 & 0.845 $\pm$ 0.063 & 0.854 $\pm$ 0.065 & 0.922 $\pm$ 0.042 \\
&   Emotion-only
& 0.466 $\pm$ 0.030 & 0.512 $\pm$ 0.014 & 0.544 $\pm$ 0.057
& 0.250 $\pm$ 0.055 & 0.505 $\pm$ 0.007 & 0.580 $\pm$ 0.155 \\
& Multimodal (no aug)
& 0.845 $\pm$ 0.038 & 0.841 $\pm$ 0.031 & 0.920 $\pm$ 0.031 & 0.867 $\pm$ 0.036 & 0.878 $\pm$ 0.037 & 0.925 $\pm$ 0.025 \\
& \textbf{Multimodal (+VAE)}
& \textbf{0.852 $\pm$ 0.038} & \textbf{0.853 $\pm$ 0.038} & \textbf{0.927 $\pm$ 0.022} & \textbf{0.876 $\pm$ 0.036} & \textbf{0.880 $\pm$ 0.034} & \textbf{0.933 $\pm$ 0.024} \\
\midrule
\multirow{3}{*}{LSTM}
& Pose-only
& 0.835 $\pm$ 0.033 & 0.827 $\pm$ 0.034 & 0.926 $\pm$ 0.019 & 0.869 $\pm$ 0.036 & 0.881 $\pm$ 0.029 & \textbf{0.935 $\pm$ 0.019} \\
& Emotion-only
& 0.407 $\pm$ 0.043 & 0.500 $\pm$ 0.000 & 0.602 $\pm$ 0.013
& 0.544 $\pm$ 0.143 & 0.599 $\pm$ 0.077 & 0.682 $\pm$ 0.030 \\
& Multimodal (no aug)
& 0.838 $\pm$ 0.043 & 0.834 $\pm$ 0.049 & 0.920 $\pm$ 0.026 & 0.874 $\pm$ 0.035 & 0.880 $\pm$ 0.039 & 0.933 $\pm$ 0.024 \\
& \textbf{Multimodal (+VAE)}
& \textbf{0.857 $\pm$ 0.036} & \textbf{0.858 $\pm$ 0.037} & \textbf{0.928 $\pm$ 0.020} & \textbf{0.876 $\pm$ 0.035} & \textbf{0.882 $\pm$ 0.037} & 0.934 $\pm$ 0.022 \\
\midrule
\multirow{3}{*}{Transformer }
& Pose-only
& 0.880 $\pm$ 0.014 & 0.879 $\pm$ 0.014 & 0.947 $\pm$ 0.015 & 0.897 $\pm$ 0.020 & 0.903 $\pm$ 0.018 & 0.948 $\pm$ 0.016 \\
& Emotion-only
& 0.577 $\pm$ 0.043 & 0.589 $\pm$ 0.008 & 0.676 $\pm$ 0.019
& 0.585 $\pm$ 0.050 & 0.636 $\pm$ 0.012 & 0.700 $\pm$ 0.014\\
& Multimodal (no aug)
& 0.881 $\pm$ 0.025 & 0.887 $\pm$ 0.027 & 0.944 $\pm$ 0.018 & 0.897 $\pm$ 0.026 & \textbf{0.908 $\pm$ 0.025} & 0.946 $\pm$ 0.019 \\
& \textbf{Multimodal (+VAE)}
& \textbf{0.888 $\pm$ 0.021} & \textbf{0.895 $\pm$ 0.017} & \textbf{0.95 $\pm$ 0.015} & \textbf{0.899 $\pm$ 0.021} & 0.905 $\pm$ 0.027 & \textbf{0.951 $\pm$ 0.017} \\
\bottomrule
\vspace{-2em}
\end{tabular}
\end{table*}

\subsubsection{Evaluation strategy and metrics} 

All experiments utilize our collected dataset, featuring three distinct indoor environments: Env1, Env2, and Env3 as shown in Table \ref{tab:dataset_summary}. Models are trained and validated with \emph{5-fold stratified cross-validation} on Env~1+2: sequences are partitioned into folds so that all frames from a sequence remain in the same split. To assess generalization to multi-person scenes, we keep Env.~3 entirely unseen during training. We perform a two-split repeated held-out protocol: Env.~1+2 are partitioned into two disjoint, stratified halves (50–50) by subject and session. We train a model on each half and evaluate both models on the same Env.~3 test set.

We report metrics at both the \textit{frame} and \textit{sequence window} levels (i.e., frame-level classification and sequence window-level classification). Frame-level uses per-frame probabilities, while sequence-level makes a decision about the human intention to interact once the model’s predicted intent probability exceeds a threshold $\tau$ for at least $k_{\text{run}}{=}7$ consecutive frames within each 15-frame sequence window (i.e., for half of the frames within each window). For ground-truth sequence labels, a window is positive when the onset annotation yields $\geq 7$ positive frames within the window; otherwise, it is negative.  At the sequence level, true positives are interacting sequences with scores \(\ge \tau\); true negatives are non-interacting sequences with scores \(< \tau\); false positives are non-interacting sequences with scores \(\ge \tau\); and false negatives are interacting sequences with scores \(< \tau\).

We report the area under the receiver operating characteristic curve (AUROC), macro-$F_1$, and balanced accuracy. Macro-$F_1$ (unweighted mean of per-class $F_1$) treats the rare \emph{intent} and common \emph{no-intent} classes equally, making it appropriate under extreme imbalance;  balanced accuracy compensates for skewed class priors and is more interpretable than raw accuracy in our setting; AUROC summarizes separability across all thresholds. All results are reported as \emph{mean $\pm$ standard deviation} across the cross-validation (CV) folds. 

\subsection{Results}

Table~\ref{tab:id} reports performance under four ablations: (a) \emph{pose-only}; (b) \emph{emotion-only}; (c) \emph{pose+emotion} without MINT-RVAE rebalancing; and (d) the full model with \emph{pose+emotion} and MINT-RVAE re-balancing. Table~\ref{tab:env3_results} similarly summarizes results on the out-of-sample Env.~3 multi-person test set. Metrics are reported at both \emph{frame} and \emph{sequence} resolutions. Three consistent findings emerge: (i) our proposed MINT-RVAE augmentation yields significant boosts in model performance across the three backbones; (ii) multimodal inputs (pose+emotion) outperform single-modality inputs; and (iii) the \texttt{Transformer} with MINT-RVAE achieves the strongest results overall.

\begin{figure*}[htbp]
    \setlength{\abovecaptionskip}{2pt}   
    \setlength{\belowcaptionskip}{0pt}   
    \centering
    \begin{subfigure}[t]{0.48\linewidth}
        \centering
        \includegraphics[width=0.85\linewidth]{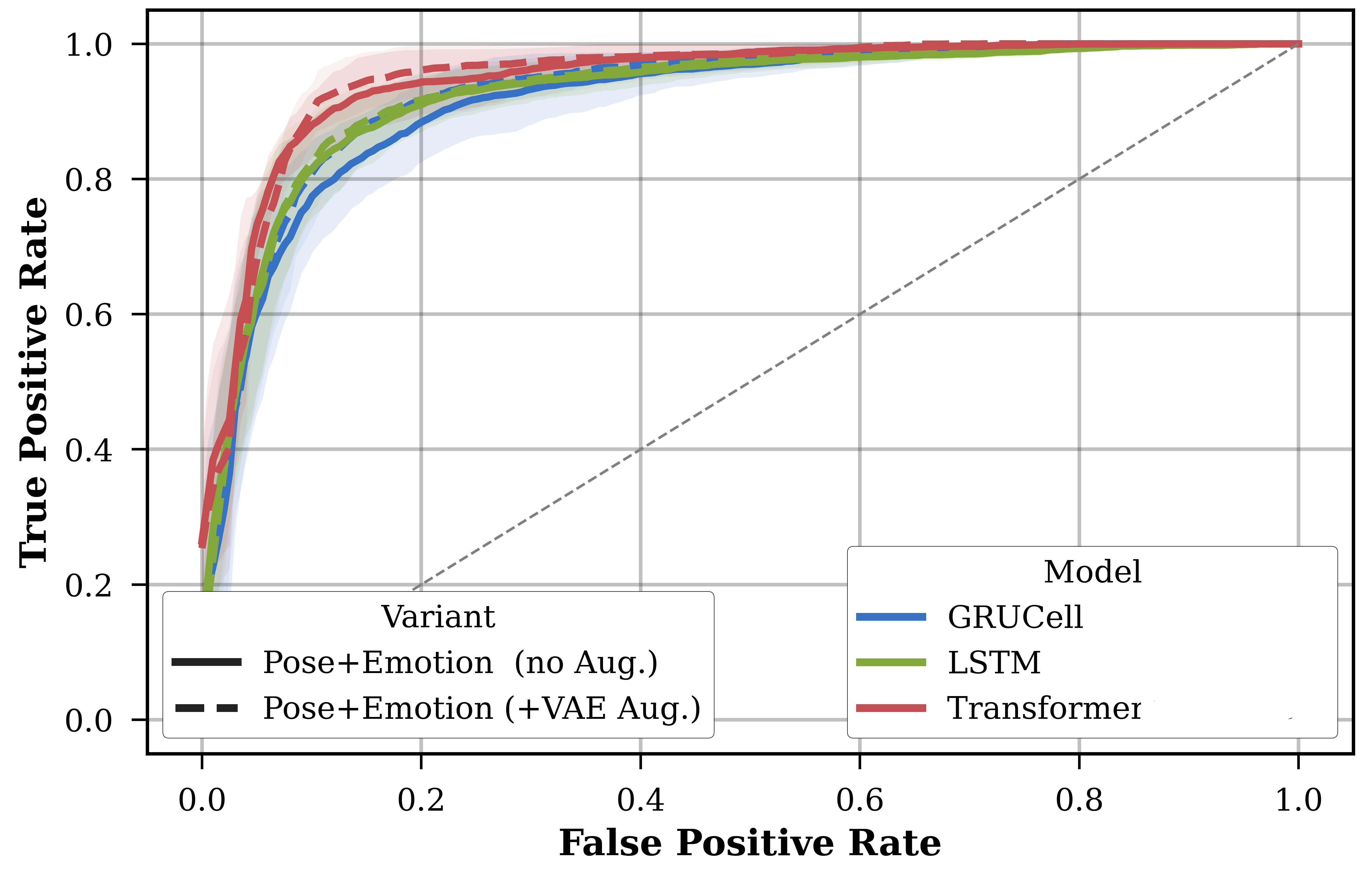}
        \label{fig:frameROC_all}
    \end{subfigure}
    \hfill
    \begin{subfigure}[t]{0.48\linewidth}
        \centering
        \includegraphics[width=0.85\linewidth]{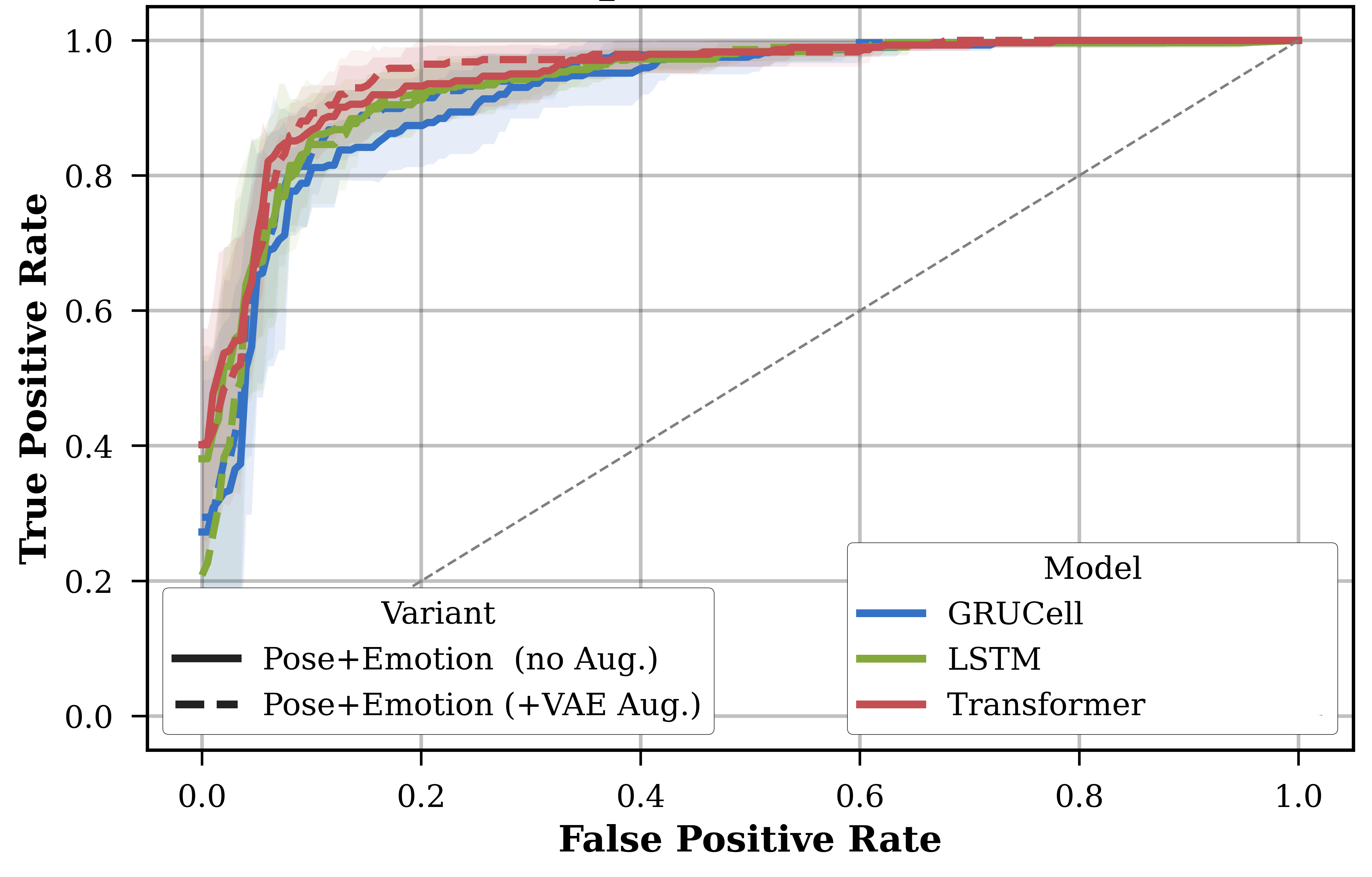}
        \label{fig:Sequence-level ROC curves}
    \end{subfigure}
    \caption{
    \textit{\textbf{ROC curves}: (a) Frame-level ROC curves; 
    (b) The sequence-level AUROC. Mean over 5-fold CV (shaded s.d.).}}
    \label{fig:roc_curves}
\end{figure*}

\begin{figure*}[htbp]
    \setlength{\abovecaptionskip}{0pt}   
    \setlength{\belowcaptionskip}{0pt}   
    \centering
    \includegraphics[width=0.90\textwidth]{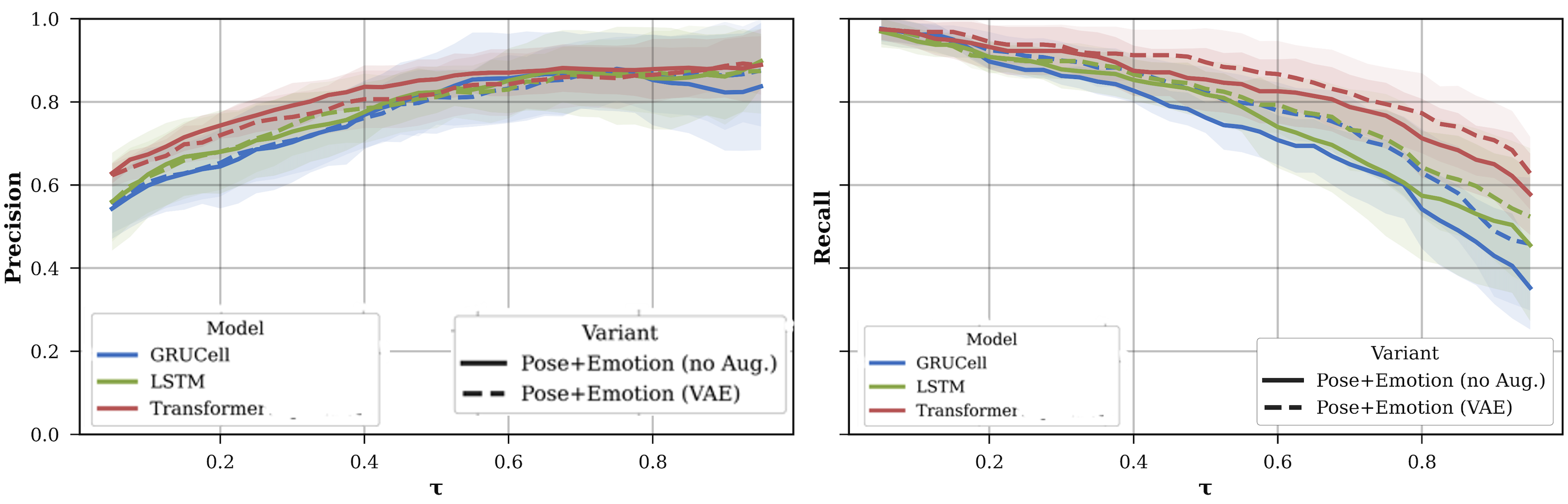}
 \caption{\textit{\textbf{Precision/Recall as a function of the decision threshold $\tau_{ao}$ for the pose+emotion models with and without MINT-RVAE augmentation}.  Mean over 5-fold CV (shaded s.d.). MINT--RVAE shifts the precision–recall trade-off upward.}}
 

    \label{fig:precision-recall}
\end{figure*} 

\subsubsection{Frame-Level assessment}

On Env.~1+2 (5-fold CV), as shown in Table~\ref{tab:id},  combining pose and emotion with MINT-RVAE  augmentation boosts the model performances across the overwhelming number of entries. On the other hand, using emotion-only input data leads to consistently weak model performances. This indicates that facial dynamics alone are \textit{insufficient} for early intent prediction, underscoring that full-body dynamics carry the \textit{primary} information about human intent, with emotion cues yielding clear benefits as a \textit{secondary} source of information when fused with the pose skeleton data. On average, our proposed \textit{Multimodal + MINT-RVAE} approach leads to a reliable gain of $\sim4\%$  across all frame-level performance indicators, compared to the pose-only baseline.

\begin{table*}[h]
\centering
\caption{\textit{\textbf{Model generalization on out-of-sample multi-person test data (Env3).} The performance is reported as the \texttt{mean} $\pm$ the \texttt{standard deviation} across the different training data folds (keeping the whole of Env3 as the test set during the evaluation procedure). Best values are indicated in bold.}}
\label{tab:env3_results}
\setlength{\tabcolsep}{3.6pt}
\begin{tabular}{l l c c c c c c}
\toprule
\textbf{Architecture} & \textbf{Variant} &
\textbf{Frame F1 (Macro)} &
\textbf{Frame Bal. Acc.} & \textbf{Frame AUROC} &
\textbf{Seq F1 (Macro)} & \textbf{Seq Bal. Acc.} & \textbf{Seq AUROC} \\
\midrule
\multirow{2}{*}{GRU}
& Multimodal (no aug) 
& 0.497 $\pm$ 0.075 & 0.540 $\pm$ 0.042 & 0.575 $\pm$ 0.247 
& 0.459 $\pm$ 0.322 & 0.610 $\pm$ 0.436 & 0.636 $\pm$ 0.361 \\
& \textbf{Multimodal (+VAE)} 
& \textbf{0.728 $\pm$ 0.020} & \textbf{0.723 $\pm$ 0.006} & \textbf{0.770 $\pm$ 0.007} 
& \textbf{0.636 $\pm$ 0.063} & \textbf{0.859 $\pm$ 0.044} & \textbf{0.810 $\pm$ 0.054} \\
\midrule

\multirow{2}{*}{LSTM}
& Multimodal (no aug) & 0.508 $\pm$ 0.130 & 0.554 $\pm$ 0.060 & 0.525 $\pm$ 0.101 & 0.425 $\pm$ 0.246 & 0.567 $\pm$ 0.084 & 0.579 $\pm$ 0.198 \\
& \textbf{Multimodal (+VAE)} & \textbf{0.742 $\pm$ 0.032} & \textbf{0.734 $\pm$ 0.026} & \textbf{0.820 $\pm$ 0.059} & \textbf{0.781 $\pm$ 0.024} & \textbf{0.810 $\pm$ 0.024} & \textbf{0.856 $\pm$ 0.026} \\
\midrule
\multirow{2}{*}{Transformer}
& Multimodal (no aug) 
& 0.749 $\pm$ 0.112 & 0.735 $\pm$ 0.097 & 0.856 $\pm$ 0.113 
& 0.812 $\pm$ 0.089 & 0.824 $\pm$ 0.036 & 0.876 $\pm$ 0.101 \\
& \textbf{Multimodal (+VAE)} 
& \textbf{0.820 $\pm$ 0.012} & \textbf{0.809 $\pm$ 0.023} & \textbf{0.932 $\pm$ 0.007} 
& \textbf{0.862 $\pm$ 0.023} & \textbf{0.869 $\pm$ 0.013} & \textbf{0.957 $\pm$ 0.002} \\

\bottomrule
\end{tabular}
\end{table*}

\subsubsection{Sequence-Level assessment}
\label{sec:results_sequence}


Crucially, as shown in Table~\ref{tab:id}, sequence-level intent prediction together with the multimodal \texttt{Transformer} model with MINT-RVAE augmentation leads to the highest reported performance (AUROC of $0.951$). This is expected since \textit{sequence-level} prediction enables the model to \textit{aggregate beliefs} from multiple consecutive time steps, increasing detection accuracy over \textit{frame-level} prediction (but this naturally comes at the cost of an increased detection latency). Fig.~\ref{fig:precision-recall} further shows that MINT--RVAE shifts the precision–recall trade-off upward, enabling higher recall at comparable precision (most notably for the \texttt{Transformer}), thus detecting more true interactions without increasing false positives.

\begin{figure*}[htbp]
    \centering
    \begin{subfigure}[t]{0.33\linewidth}
        \centering
        \includegraphics[width=\linewidth]{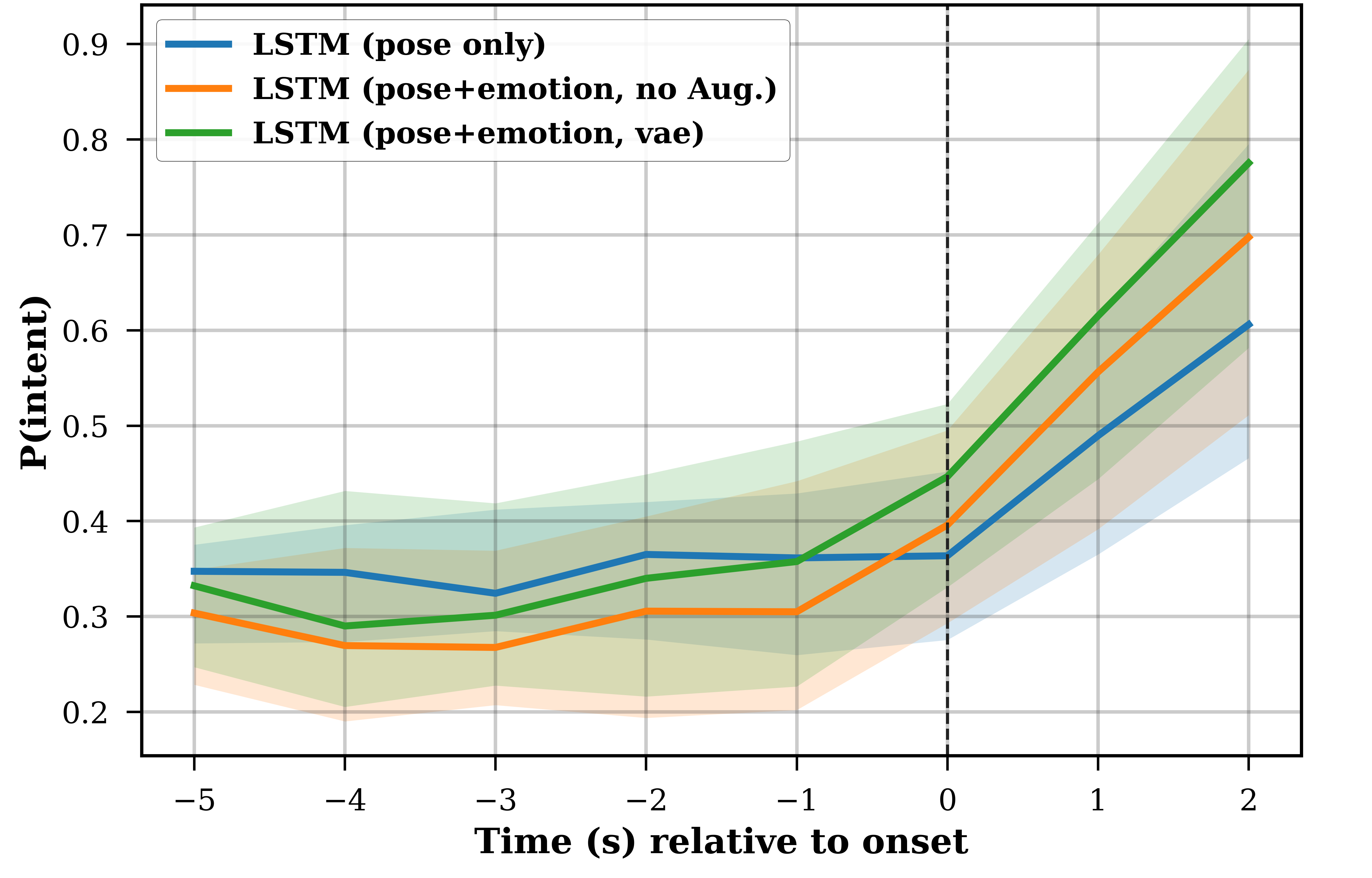}
        \caption{LSTM}
    \end{subfigure}
    \hspace{0.001\linewidth}
    \begin{subfigure}[t]{0.32\linewidth}
        \centering
        \includegraphics[width=\linewidth]{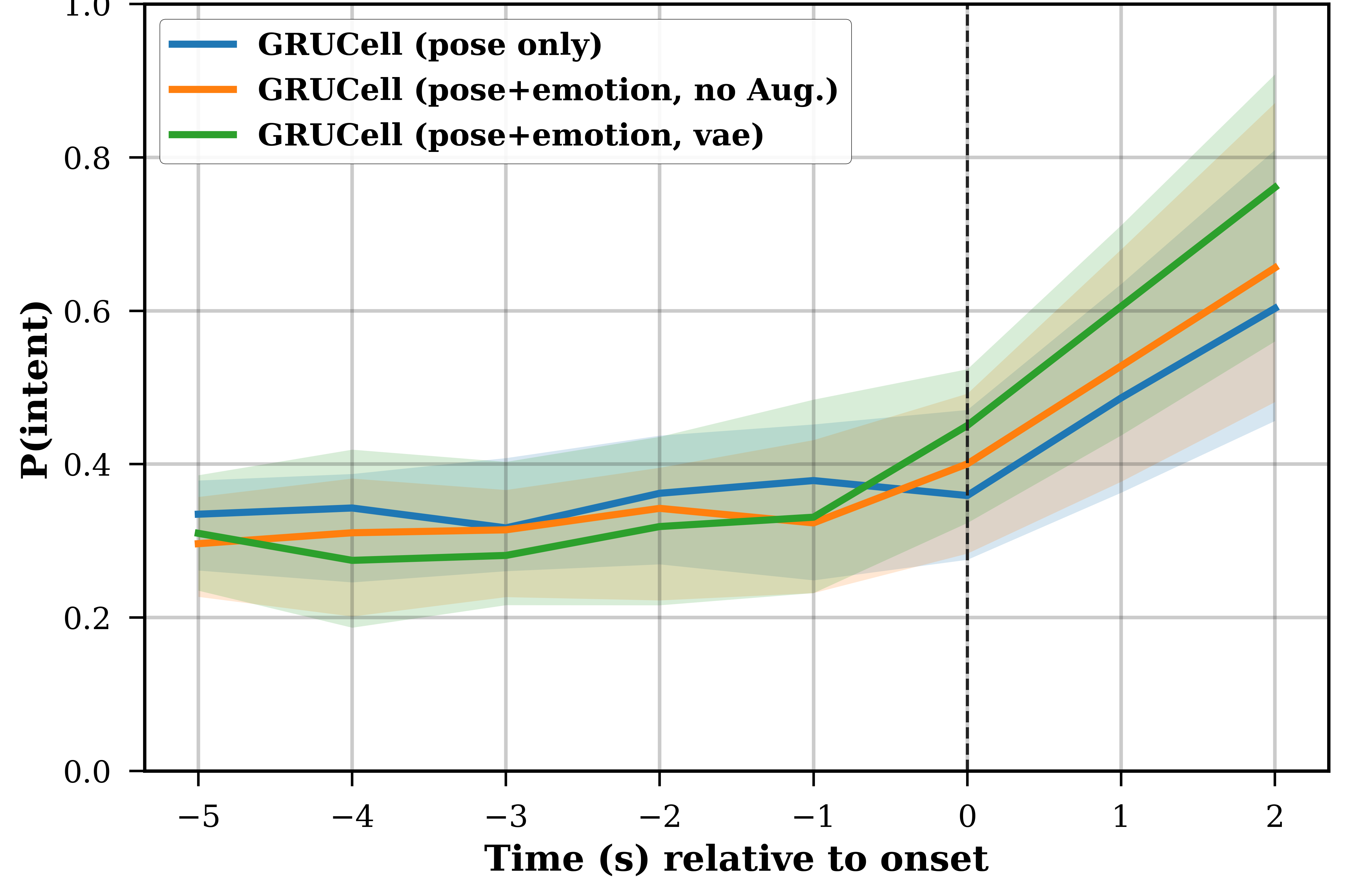}
        \caption{GRU}
    \end{subfigure}
    \hspace{0.001\linewidth}
    \begin{subfigure}[t]{0.32\linewidth}
        \centering
        \includegraphics[width=\linewidth]{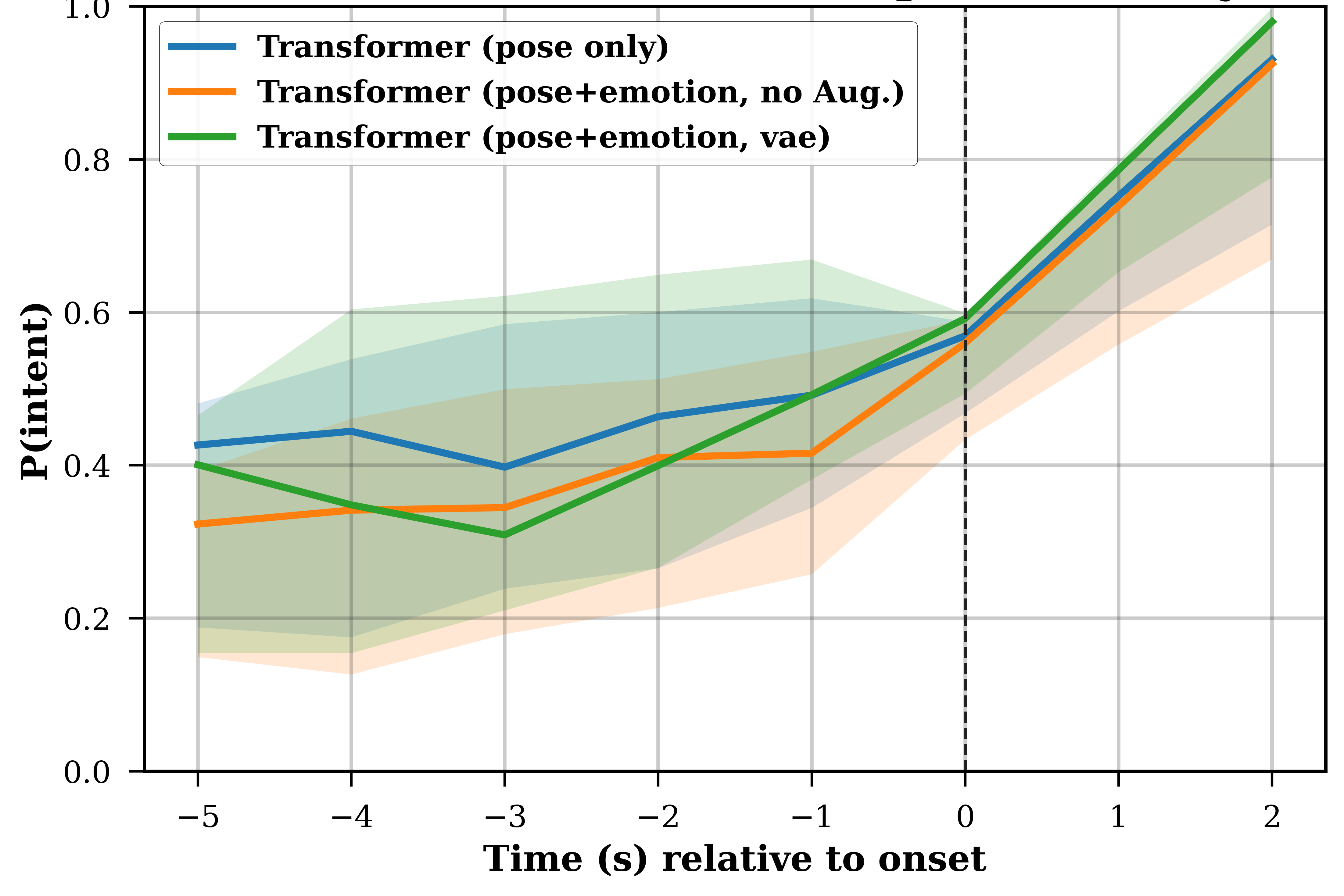}
        \caption{Transformer}
    \end{subfigure}
    \caption{\textit{\textbf{Median predicted probability of interaction over time}, aligned to the annotated intent onset (\(t=0\)). For each positive sequence, \(t=0\) marks the first frame at which the subject intends to interact.}}

    \vspace{-1em}
    \label{fig:prob_vs_time}
\end{figure*}

We further analyze the temporal evolution of the human interaction intent probability returned by the models relative to the annotated label onset (i.e., the time step when a human decides to interact with the robot). Fig.~\ref {fig:prob_vs_time}  shows positive sequences temporally aligned at interaction onset (\(t{=}0\)). Across models, scores increase as the onset approaches and surge thereafter, confirming temporal evidence aggregation. Adding emotion features and MINT--RVAE augmentation (green) \textit{systematically} elevates the trajectories and steepens the pre-onset slope relative to pose-only (blue) and multimodal without augmentation (orange). The effect is most pronounced for the \texttt{Transformer} and \texttt{GRU}, which exhibit earlier separation from the baselines before onset and higher confidence immediately after onset. The \texttt{LSTM} shows a smaller but consistent gain. Practically, these shifts indicate earlier and more reliable intent detection at comparable variability, enabling shorter reaction time.

\begin{table*}[t]
\centering
\small
\setlength{\tabcolsep}{2pt} 
\renewcommand{\arraystretch}{1.0} 
\caption{\textit{\textbf{Comparison against recent intent-prediction methods in HRI.}}}
\label{tab:comparison}
\begin{tabularx}{\linewidth}{@{} l l c c c X @{}}
\toprule
Work & Sensors & Resolution & Features & Data Re-balancing & Performance  \\
\midrule
Trick et al. (2023) \cite{trick2023can} & RGB--D + mic. & Sequence-level. & Pose, gaze, speech & No & $F_{1}\!\approx\!0.81$ \\
\midrule
Abbate et al. (2024) \cite{abbate2024self} & RGB--D & Sequence-level & Pose & No & AUROC $\approx$ 0.90 \\
\midrule
Arreghini et al. (2024) \cite{arreghini2024predicting} & RGB--D + gaze & Sequence-level & Pose, gaze & No & AUROC: 0.91.2 \\
\midrule
\textbf{This work} & \textbf{RGB-only} & \textbf{Frame + Sequence-level} & \textbf{Pose, emotion} & \textbf{Yes} & \textbf{AUROC: $0.951\!\pm\!0.017$} \\
\bottomrule
\end{tabularx}
\end{table*}

\subsection{Model generalization to the multi-person setting}
\label{generalization}

Table~\ref{tab:env3_results} evaluates the models trained on single-person data (Envs.~1–2) directly on the held-out two-person set (Env.~3), thus probing out-of-distribution generalization. Importantly, in these experiments, our models were not trained with any data from the more challenging two-person setting.  Table~\ref{tab:env3_results} shows that using our proposed MINT-RVAE data re-balancing method during training has a highly significant effect on the generalization of all backbones, yielding an average multiplicative gain of $\sim 25\%$ across all the performance indicators. The best performance is achieved by the \texttt{Transformer}+MINT--RVAE (sequence AUROC \(0.957\), frame AUROC \(0.932\)), while substantial \texttt{GRU} and \texttt{LSTM} gains highlight the clear generalization benefits of augmentation under multi-person scenes.

\subsection{Comparison with prior works}
Table~\ref{tab:comparison} compares our approach with recent HRI intent-prediction systems \cite{arreghini2024predicting,abbate2024self,trick2023can}. Most prior work relies on RGB--D sensing and reports sequence-level metrics, whereas ours is RGB-only with \emph{onset-accurate}, frame-level evaluation. Under our dataset and protocol, we obtain AUROC \(0.951 \pm 0.017\) (sequence) and \(0.95 \pm 0.015\) (frame). In contrast, acknowledging dataset and protocol differences, Abbate \textit{et al.} \cite{abbate2024self} report AUROC just above 0.9 using an RGB--D self-supervised pose model, and Arreghini \textit{et al.} \cite{arreghini2024predicting} report AUROC \(=0.912\) with gaze\,+\, pose (\(\approx 0.845\) pose-only). Their frame-wise scores replicate the sequence label across frames leading to \textit{no} onset annotation. Trick \textit{et al.} \cite{trick2023can} evaluate event-level detection (counting an intention as detected if any frame within the window is positive) and report \(F_{1}\!\approx\!0.81\) (rather than reporting the more robust AUROC metric). 

Now, when it comes to class imbalance, interaction positives are scarce in public deployments \cite{thompson2024pard,natori2025mall}, and several baselines either do not re-balance, or emphasize accuracy-style metrics that can be misleading under class imbalance. Our training incorporates sequence-level augmentation (MINT-RVAE), and we report both macro-\(F_{1}\) and balanced accuracy alongside AUROC. Overall, our methods report competitive or superior performance in terms of AUROC while: \textit{i)} using a single RGB sensor, \textit{ii)} providing true frame-level onset evaluation, and \textit{iii)} explicitly addressing the class imbalance problem, leading to superior out-of-sample model generalization for real-world deployments.

\section{Conclusion}
\label{concsection}
This study presents a novel pipeline for detecting the intent of human subjects for interacting with robots with \textit{frame-level} precision using only RGB camera data. This is in contrast to most prior studies, which use less cost-effective multi-sensor setups (e.g., RGB-D) and provide sequence-level predictions, degrading the robot reaction time and, consequently, the general quality of service. Furthermore, to alleviate the well-known data imbalance issues encountered in most HRI datasets, this study also presented the design of a multimodal recurrent VAE, termed \textit{MINT-RVAE}, specially tailored for the synthetic generation of highly realistic HRI data. On the held-out two-person challenging dataset, MINT--RVAE improves generalization markedly for all models. Furthermore, our numerous experiments have demonstrated that our approach matches or surpasses the recent baselines while using only RGB, achieving CV sequence AUROC \(0.951 \pm 0.017\) and frame AUROC \(0.95 \pm 0.015\)---exceeding reported RGB-D results (\(\approx 0.90\)--\(0.912\) AUROC)---and doing so with true \textit{onset} frame-level decisions. Finally, we openly released our newly acquired dataset featuring frame-level annotations with the hope of benefiting future research.

\bibliographystyle{IEEEtran}
\bibliography{references}
\end{document}